\pdfoutput=1 
\documentclass[letterpaper]{article} 
\usepackage{aaai2026}  
\usepackage{times}  
\usepackage{helvet}  
\usepackage{courier}  
\usepackage[hyphens]{url}  
\usepackage{graphicx} 
\usepackage{natbib}  
\usepackage{caption} 
\usepackage{algorithm}
\usepackage{algorithmic}

\usepackage[capitalize]{cleveref}
\usepackage{booktabs}
\usepackage{multicol}
\usepackage{multirow}
\usepackage{makecell}
\usepackage{amssymb}
\usepackage{pifont}
\usepackage{xcolor}
\newcommand{\etal}{\textit{et al.}}
\newcommand{\vs}{\textit{vs\ }}

\usepackage{newfloat}
\usepackage{listings}
\DeclareCaptionStyle{ruled}{labelfont=normalfont,labelsep=colon,strut=off} 
\floatstyle{ruled}
\newfloat{listing}{tb}{lst}{}
\floatname{listing}{Listing}
\title{CPCL: Cross-Modal Prototypical Contrastive Learning for \\Weakly Supervised Text-based Person Retrieval\thanks{This paper is currently under peer review.}}
\author{
    Xinpeng Zhao\textsuperscript{\rm 1},
    Yanwei Zheng\textsuperscript{\rm 1},
    Chuanlin Lan\textsuperscript{\rm 1}\\
    Xiaowei Zhang\textsuperscript{\rm 2},
    Bowen Huang\textsuperscript{\rm 1},
    Jibin Yang\textsuperscript{\rm 3},
    Dongxiao Yu\textsuperscript{\rm 1}
}
\affiliations{
    \textsuperscript{\rm 1}School of Computer Science and Technology, Shandong University,
    \textsuperscript{\rm 2}College of Computer Science and Technology, Qingdao University,
    \textsuperscript{\rm 3}Research Center for Mathematics and Interdisciplinary Science, Shandong University.
    
}

\usepackage{bibentry}

\begin{document}

\maketitle

\begin{abstract}
Weakly supervised text-based person retrieval seeks to retrieve images of a target person using textual descriptions, without relying on identity annotations and is more challenging and practical.
The primary challenge is the intra-class differences, encompassing \textit{intra-modal feature variations} and \textit{cross-modal semantic gaps}.
Prior works have focused on instance-level samples and ignored prototypical features of each person which are intrinsic and invariant.
Toward this, we propose a Cross-Modal Prototypical Contrastive Learning (CPCL) method.
In practice, the CPCL introduces the CLIP model to weakly supervised text-based person retrieval to map visual and textual instances into a shared latent space.
Subsequently, the proposed Prototypical Multi-modal Memory (PMM) module captures associations between heterogeneous modalities of image-text pairs belonging to the same person through the Hybrid Cross-modal Matching (HCM) module in a many-to-many mapping fashion.
Moreover, the Outlier Pseudo Label Mining (OPLM) module further distinguishes valuable outlier samples from each modality, enhancing the creation of more reliable clusters by mining implicit relationships between image-text pairs.
We conduct extensive experiments on popular benchmarks of weakly supervised text-based person retrieval, which validate the effectiveness, generalizability of CPCL.

\end{abstract}

\begin{figure}[ht]
    \centering{
    \includegraphics[scale=1.1]{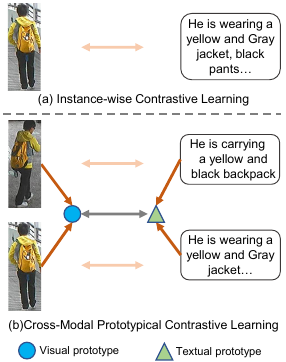}}
    \caption{Illustration of instance-wise contrastive learning and cross-modal prototypical contrastive learning.
    (a): It 
    treats two samples as a negative pair as long as they are from different pseudo labels, regardless of their semantic similarity, 
    (b): The prototype acts as a bridge, overcoming intra-modal feature variations and mitigating the cross-modal semantic gaps. 
    }
    \label{fig:motivation}
\end{figure}

\section{Introduction}

Text-based person retrieval is designed to retrieve all images of a pedestrian across multiple cameras based on a textual description\cite{CFine, IRRA, bai2023rasa, cheng2024ACCV}. 
A widely adopted framework for text-based person retrieval involves projecting the image and text input to a common latent space through separate encoders and aligning the features of the same identity in the latent space. 
The annotation of the identity is crucial to mitigate the variations \textit{intra-class} in the alignment process.
On the one hand, the annotation helps mitigate \textit{intra-modal feature variations}.
For example, images of a person could be captured by different cameras with varying illumination, view, and background. 
The annotations provide supervisory information to address intra-modal variation effectively.
On the other hand, identity labels narrow down the \textit{cross-modal semantic gaps} between images and texts. 
For instance, a textual description should be matched not only with its paired image but also with other images of the same person.
However, obtaining identity annotations for the dataset is costly and time consuming. 
In a more practical setting, the weakly supervised text-based person retrieval \cite{CMMT, CAIBC, gong2024enhancing} is considered, where the training set only contains text-image pairs without any identity annotations.

The absence of identity labels consequently introduces intra-class variations at both cross-modal and intra-modal levels. 
To address the these challenges, we propose a Cross-Modal Prototypical Contrastive Learning (CPCL) method to guide the multi-modal leaning, inspired by the human cognitive process\cite{ross1999prototype,lakoff2007cognitive} and the progress in other visual tasks\cite{snell2017prototypical,PCL}.
Intuitively, humans have the metacognitive ability to access knowledge to classify a sample as a certain type and then use their knowledge of this type to determine how to deal with it\cite{fleming2012neural}.
We propose a noval learning scheme where the visual samples of each person are assigned a visual prototype to leverage this cognitive behavior mechanism.
Similarly, textual samples are assigned a textual prototype.
Each visual and textual instance is assigned a prototype that corresponds to its pseudo label, which helps the model to learn intra-modal discriminative representations while overcoming intra-modal feature variations while mitigating the cross-modal semantic gaps, as illustrated in \cref{fig:motivation}(b).

From a technical standpoint, the prototype representations are closely aligned with the class centers obtained via clustering algorithms.
To obtain good initial prototype representations, we cluster images and texts separately to get a better initialization of the prototypes and assign each sample a pseudo-label using the DBSCAN algorithm\cite{DBSCAN}.
Considering that the prototypical representations are constantly changing as the model updates, we propose the Prototypical Multi-modal Memory (PMM) module, which dynamically maintains the prototypical representations so that the prototypical representation can be kept up-to-date.
To mitigate intra-class variation, we propose a Hybrid-level Cross-modal Matching (HCM) module.
It includes the Prototypical Contrastive Matching (PCM) loss between instance representations and prototypical representations of different modalities provided by PMM, and Instance-level Cross-modal Projection Matching (ICPM) loss in each mini-batch with positive and negative samples obtained according to the pseudo labels.
During the training process, we discover a large number of un-clustered samples.
We propose an Outlier Pseudo Label Mining (OPLM) module to identify valuable outlier samples from each modality, enhancing the clustering quality by mining implicit relationships between image-text pairs.
To obtain a better-aligned latent space of visual and textual encoders, we leverage CLIP\cite{CLIP} as the backbone for our framework. 
Our main contributions can be summarized as follows:
\begin{itemize}
    \item We propose a Cross-Modal Prototypical Contrastive Learning (CPCL) framework designed to mitigate the intra-modal representation variations and cross-modal semantic gaps. We conduct extensive experiments on popular benchmarks of weakly supervised text-based person retrieval, which validate the effectiveness, the generalizability of CPCL.
    \item We propose a Prototypical Multi-modal Memory (PMM) for Hybrid-level Cross-modal Matching (HCM), which helps CLIP establish the associations between heterogeneous modalities of image-text pairs belonging to the same person.
    \item We introduce an Outlier Pseudo Label Mining (OPLM) module that further distinguishes valuable outlier samples by leveraging the implicit relationships between image-text pairs. 
\end{itemize}

\begin{figure*}[ht]
    \centering
    \includegraphics[scale=1.2]{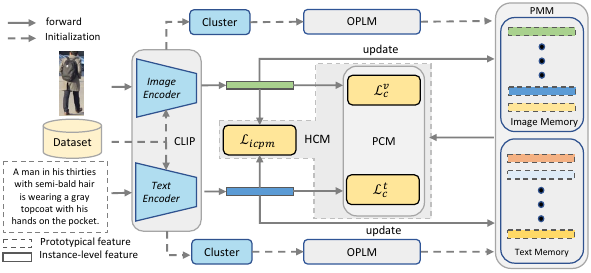}
    \caption{
    Our proposed Cross-modal Prototypical Contrastive Learning assigned each visual and textual sample a prototype corresponding to modality.
    PMM: Prototypical Multi-modal Memory module dynamically maintains the prototypical features.
    HCM: Hybrid-level Cross-modal Matching module consists of Prototypical Contrastive Matching (PCM) loss and Instance-level Cross-modal Projection Matching (ICPM) loss.
    OPLM: Outlier Pseudo Label MIning module identifies valuable un-clustered samples by mining implicit relationships between image-text pairs.
    }
    \label{fig:pipeline}
\end{figure*}

\section{Rekated Work}
\subsection{Text-based Person Retrieval}
Text-based person retrieval differs from the classic image-based person retrieval (also known as person re-identification) problem\cite{zheng2015scalable,SpCL,zhao2022cluster}, and its objective is to identify person images based on a textual description.
Text-based person retrieval was originally proposed by Li\cite{CUHK-PEDES}, which utilized a CNN-RNN network to learn cross-modal representation at the global level and provided a benchmark dataset that sparked extensive research in this domain.
As the fields of vision and language have evolved, various innovative approaches have emerged, addressing complex challenges in multimodal contexts\cite{IRRA, bai2023rasa, CFine, cheng2024ACCV}.
With increasing attention to CLIP\cite{CLIP}, more and more efforts\cite{CFine,IRRA,bai2023rasa,deng2024dualfocus,cheng2024ACCV} have endeavored to leverage CLIP's powerful knowledge to establish an effective mapping relationship between images and texts.
Leveraging labeled data, especially the id annotation, these methods achieve significantly improved performance and greatly contributing to the development of the field.
These methods belong to the field of fully supervised learning, but in the practical scenario, the ID annotation is costly and time-consuming.

\subsection{Weakly Supervised Text-based Person Retrieval}
Research on weakly supervised text-based person retrieval tasks is scarce. 
There are only a few works on weakly supervised text-based person search. 
CMMT\cite{CMMT} first proposed the task and proposed MPLR to mine valuable un-clustered samples while employing Text-IOU to guide the model in learning image-text matching and to mine challenging samples. 
Our Outlier Pseudo Label Mining (OPLM) module is similar to it, but the core difference is that OPLM considers the existence of one-to-one and one-to-many relationships between images and text, not just one-to-one relationships in dataset. 
Based on weakly supervised text-based person retrieval task, some works\cite{gong2024enhancing, du2024contrastive, gong2024cross} propose a new task, unsupervised incomplete cross-modal retrieval, where the only difference is that not all images (text) have corresponding text (images) in training dataset. 
For a fair comparison, we only choose the performance of these methods on the complete dataset. 
ECCA\cite{gong2024enhancing} proposed an enhanced prototype-wise granularity alignment module that achieves alignment of global visual and textual embeddings by utilizing cross-modal unified prototypes for both image and text modality as intermediaries.
As a comparison, it is similar to our prototypical multi-modal memory module, but there are three different points:
(1) Our prototypes are modality-specific rather than unified; that is, images and texts are each associated with their own set of prototypes. This design choice is motivated by the need to mitigate intra-modal feature variations.
(2) The pseudo labels we use are obtained by clustering rather than by the Sinkhorn-Knopp algorithm\cite{cuturi2013sinkhorn}.
(3) Finally, our prototypes are updated using a momentum approach rather than a learnable approach.
While CCL\cite{du2024contrastive} and CANC\cite{gong2024cross} propose intra-modal prototype-based contrastive learning frameworks, our approach, CPCL, extends this paradigm by incorporating both intra-modal and inter-modal prototype-based contrastive matching losses, thereby enabling more comprehensive representation learning across and within modalities.

\section{Method}
\subsection{Overview}
In this section, we will elaborate on the proposed CPCL framework.
First, we introduce the Prototpcial Multi-model Memory (PMM) moduel in \cref{section: PMM}, where we will describe the mechanism of its initialization and update.
Then, in \cref{section: HCM}, we delve into the Hybrid Cross-modal Matching (HCM) module, which encompasses the Protopypical Contrastive Matching (PCM) and Instance Cross-modal Project Matching (ICPM).
Finally, we elaborate the proposed Outlier Pseudo Lable Mining (OPLM) in \cref{section: OPLM}, which has the refined learning stage and the supplementary learning stage.
The overall framework of CPCL is illustrated in \cref{fig:pipeline}.

\subsection{Prototpcial Multi-modal Memory}\label{section: PMM}
Humans can be capable of classifying a sampler as a certain type and then use their knowledge of this type to determine how to deal with it\cite{fleming2012neural}.
The proposed cross-modal prototypical contrastive learning simulates this cognitive behavior mechanism by bringing instance-level samples closer to their corresponding prototypes.
Considering that the prototypical features are constantly updating as the training process, we propose the Prototypical Multi-modal Memory (PMM) module to dynamically maintain the prototypical features and keep them up-to-date.

\subsubsection{Memory Initialization}
Firstly, we employ the Image encoder to extract all image features, resulting in $F^v = \{f^v_1, \ldots, f^v_{n_v}\}$. 
The superscript $v$ indicates that it is a visual feature and $n_v$ denotes the number of images. 
Then we utilize a clustering algorithm to cluster all image features and obtain pseudo labels $Y^v = \{y^v_1, \ldots, y^v_{n_v}\}$.

To initialize the prototypical features for each class among the visual samples, we calculate a prototypical feature with pseudo label $i$ based on
\begin{equation}\label{eq:image-memory-init}
    c^v_i = \frac{1}{\| Y^v_i \|} \sum_{f^v \in Y^v_i} f^v,
\end{equation}
where $c^v_i$ represents the prototype feature with pseudo label $i$ in the image memory, while $Y^v_i$ denotes a set of all image samples with the pseudo label $i$. 
Furthermore, the symbol $||\cdot||$ indicates the number of elements in a set.
The initialization process for text memory follows the same procedure as image memory.

\subsubsection{Memory momentum updating}
After each training iteration, both image and text sample features within the given mini-batch are utilized to momentum update the prototypical features stored in prototypical memory, following the pseudo labels derived from clustering.
Specifically, the feature of a certain image sample within the mini-batch, denoted as $f^v_i$, we perform the update to the image memory using
\begin{equation}\label{eq:image-memory-update}
c^v_i \leftarrow m^v c^v_i + (1-m^v)f^v_i ,
\end{equation}
where $m^v$ represents the image memory momentum update coefficient, which controls the momentum update rate of class center samples. 
Similarly, For each text feature $f^t_i$ within the mini-batch, the text memory will also be momentum updated following the protocol as mentioned above. 
In our paper, we set $m^v$ and $m^t$ as 0.9 as default in our experiments.

\subsection{Hybrid Cross-modal Matching}\label{section: HCM}
\subsubsection{Protopypical Contrastive Matching}
We employ two optional Protopypical Contrastive Matching (PCM):
cross-modal PCM and single-modal PCM, denoted as $PCM_c$ and $PCM_s$, respectively.
The formulation of the $PCM_c$ loss for image-to-text direction is as follows.
\begin{equation}\label{eq:image2text-loss}
\mathcal L^{v}_{c} = 
        -\log
        \frac{ \exp \left(f^v \cdot c^{t+} / \tau^v \right)}
        {\sum_{i=1}^{N^t} \exp \left(f^v \cdot c^t_{i} / \tau^v \right)} 
\end{equation}
where $f^v$ denotes the image feature. The parameters $\tau^v$ and $N^t$ represent the learnable temperature parameter for the image memory and the number of cluster categories for captions, respectively.
Symmetrically, the $PCM_c$ loss for text-to-image direction, denoted by $\mathcal{L}^t_c$ is derived by interchanging the roles of $f^v$ with $f^t$ and $c^t$ with $c^v$ within the framework of equation \cref{eq:image2text-loss}:
\begin{equation}\label{eq:text2image-loss}
\mathcal L^{t}_{c} = 
        -\log
        \frac{ \exp \left(f^t \cdot c^{v+} / \tau^t \right)}
        {\sum_{i=1}^{N^v} \exp \left(f^t \cdot c^v_{i} / \tau^t \right)}.
\end{equation}

Take \cref{eq:image2text-loss} as an example, we utilize the image feature $f^v$ to calculate the loss in text memory. 
Specifically, based on the paired textual feature $f^t$, we can identify a positive sample $c^{t+}$ in the text memory, though maybe the cluster id of the image sample is different from the cluster of the paired caption sample. 
Moreover, we consider samples from different text memory classes as negative samples. 
This mechanism helps the model to learn the mapping relationship between images and captions globally.
Similarly, \cref{eq:text2image-loss} facilitates the model in learning the mapping relationship between text and image.
In summary, the bi-directional $PCM_c$ loss is calculated by:
\begin{equation}\label{eq:cross-modal-loss}
    \mathcal L_{pcm}^c = \mathcal L^{v}_{c} + \mathcal L^{t}_{c}  
\end{equation}

Similarly, $PCM_s$ loss $\mathcal L^{s}_{pcm}$ is also bi-directional and composed of $\mathcal L^{v}_s$ and $\mathcal L^{t}_s$:
\begin{equation}\label{eq:image2image-loss}
\mathcal L^{v}_{s} = 
        -\log
        \frac{ \exp \left(f^v \cdot c^{v+} / \tau^v \right)}
        {\sum_{i=1}^{N^t} \exp \left(f^v \cdot c^v_{i} / \tau^v \right)} 
\end{equation}
and
\begin{equation}\label{eq:text2text-loss}
\mathcal L^{t}_{s} = 
        -\log
        \frac{ \exp \left(f^t \cdot c^{t+} / \tau^t \right)}
        {\sum_{i=1}^{N^v} \exp \left(f^t \cdot c^t_{i} / \tau^t \right)}.
\end{equation}
Similarly, the $PCM_s$ loss is calculated by: 
\begin{equation}\label{eq:single-modal-loss}
    \mathcal L_{pcm}^s = \mathcal L^{v}_{s} + \mathcal L^{t}_{s}  
\end{equation}

In our paper, cross-modal PCM is adopted by default, $\mathcal L_{pcm} = \mathcal L_{pcm}^c$.
It is essential to emphasize that the integration of PCM and PMM is imperative for their collaborative functionality. 
These two components must be combined to work better.

\subsubsection{Instance Cross-modal Project Matching}
In addition to matching at the prototypical level, instance-level matching is also necessary.
Inspired by the CMPM\cite{CMPM}, we applied Instance Cross-modal Project Matching (ICPM) loss for alignment at the instance level.
Specifically, we calculate the probability of cross-modal matching pairs by the following softmax function:
\begin{equation}\label{eq:icpm-pij}
p^{v2t}_{i,j} = 
    \frac{ \exp \left(f^v_i \cdot f^t_j / \tau \right)}
    {\sum_{k=1}^{N} \exp \left(f^v_i \cdot f^t_{k} / \tau \right)},
\end{equation}
where footnote $i$ and $j$ denote the indexes of features in a mini-batch. 
$N$ is the number of image-text pairs.
The pseudo cross-modal matching probability is  $q^v_{i,j} = y_{i,j} /\sum_{k=1}^{N}y_{i,k}$.
$y_{i,j} = 1$ means that $(f^v_i, f^t_j)$ is a matched pair from the same pseudo label, while $y_{i,j} = 0$ indicates the unmatched pair. 
Then the ICPM loss is computed by
\begin{equation}\label{eq:icpm-img-loss}
    \mathcal{L}_{v2t}^{icpm} = KL(\mathbf{p^v_i}\| \mathbf{q^v_i}) = \frac{1}{N} \sum_{i=1}^{N}\sum_{j=1}^{N}p^v_{i,j}\log(\frac{p^v_{i,j}}{q^v_{i,j} + \epsilon}),
\end{equation}
where $\epsilon$ is a small number to avoid numerical problems.
Symmetrically, the ICPM loss from text to image $\mathcal{L}_{t2i}^{icpm}$ can be formulated by exchanging $f^v$ and $f^t$ in \cref{eq:icpm-pij,eq:icpm-img-loss}:
\begin{equation}\label{eq:icpm-pijt}
p^{t2v}_{i,j} = 
    \frac{ \exp \left(f^t_i \cdot f^v_j / \tau \right)}
    {\sum_{k=1}^{N} \exp \left(f^t_i \cdot f^v_{k} / \tau \right)},
\end{equation}
and 
\begin{equation}\label{eq:icpm-text-loss}
    \mathcal{L}_{t2v}^{icpm} = KL(\mathbf{p^t_i}\| \mathbf{q^t_i}) = \frac{1}{N} \sum_{i=1}^{N}\sum_{j=1}^{N}p^t_{i,j}\log(\frac{p^t_{i,j}}{q^t_{i,j} + \epsilon}).
\end{equation}
The bi-directional ICPM loss is calculated by:
\begin{equation}
  \mathcal{L}_{icpm} = \mathcal{L}_{v2t}^{icpm} + \mathcal{L}_{t2v}^{icpm}. 
\end{equation}

\textbf{Overall Loss}
By combining the losses defined above, the final objective of CPCL is formulated as:
\begin{equation}\label{eq:overall-loss}
\mathcal L_{overall} = \mathcal L_{pcm} + \mathcal L_{icpm}
\end{equation}

\begin{figure}[t!]
    \centering
    \includegraphics[scale=1.0]{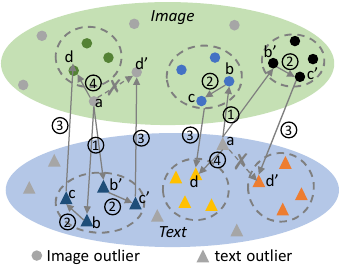}
    \caption{
    Illustration of Outlier Pseudo Label Mining (OPLM). 
    Different colors indicate different cluster IDs.
    \ding{192}: For an outlier instance, group all the paired and clustered samples as a set B\{b, b'\}.
    \ding{193}: Identify the nearest same modality samples set to B and denote it as C\{c, c'\}.
    \ding{194}: Search the set of paired instances D \{d, d'\} of C
    in the other modality and remove any outlier samples from D \{d, d'\}. 
    \ding{195}: Assign the closest sample's cluster ID to the outlier instance.
    }
    \label{fig:OPLM}
\end{figure}

\subsection{Outlier Pseudo Label Mining}\label{section: OPLM}
We propose an Outlier Pseudo Label Mining (OPLM) module that leverages existing image-text pairs information to mine valuable outlier samples, as illustrated as \cref{fig:OPLM}. 
OPLM consists of two stages: the refined learning stage and the supplementary learning stage, respectively.

\subsubsection{The Refined Learning Stage}
At the refined learning stage, the OPLM algorithm needs to be applied separately to both visual and textual modalities to identify valuable outlier samples from each modality.
Consequently, it can be divided into two distinct components: $OPLM_{v \rightarrow t}$ and $OPLM_{t \rightarrow v}$.

For example, in $OPLM_{v \rightarrow t} $, for an outlier image sample $v^o_i $, we can find the corresponding text description according to image text corresponding index.
However, it should be noted that there may be more than one paired text for one image.
Therefore, its paired captions are 
\begin{equation}\label{eq:oplm-1}
P^t  = PIS(v^o_i)\ ,
\end{equation}
where $P^t =\{t_1,\ldots, t_p\}$ represents that there are $p$ captions paired with this image, and $PIS(\cdot) $ denotes the process of paired instance searching in text modality. 
Subsequently, we filter $P^t$ by eliminating outlier text samples and obtain $P^t_{filter}$. 
If $||P^t_{filter}|| = 0$, $v^o_i$ remains as an outlier image sample. 
In contrast, if $P^t_{filter} = \{t_1,\ldots, t_k\}$, it indicates that there are $k$ clustered samples. 
We then process all samples in the collection $P^t_{filter} $ to identify their corresponding images:
\begin{equation}\label{eq:oplm-2}
P^v = \{ PIS(t_i) \mid t_i \in P^t_{filter} \ , i=1,\ldots,k \}\ ,
\end{equation}
where, $P^v = \{v_1,\ldots , v_k \}$ denotes $k$ images corresponding to $P^t_{filter} $ in a one-to-one manner. 
Following this, $P^v$ undergoes a filtering process in which outlier image samples are removed, resulting in $P^v_{filter}$. 
If $||P^v_{filter}|| = 0 $, then $v ^ o_i $ is kept as an outlier image sample. 
Subsequently, the distance from $v^o_i $ to all samples in $P^v_{filter} $ is calculated, and the nearest sample is identified as $v_i $
\begin{equation}\label{eq:oplm-3}
v_i = \mathop{\arg \max}\limits_{v_i \in P^v_{filter}, v_i \not \in U^v}
\langle v^o_i \cdot v_i \rangle, i=1, \ldots,q \ ,
\end{equation}
where $\langle \cdot ,\cdot \rangle$ denotes the inner product, and $U ^ v $ is a set and is initialized as empty which is used to collect excluded samples.
If the pseudo label of $v_i$ is denoted as $k$, then $v^o_i $ is added to the collection $C^v_k$, which represents the set of all visual instances belonging to pseudo label $k$
\begin{equation}\label{eq:oplm-4}
C^v_k \leftarrow \{C^v_k, v^o_i\}\ ,
\end{equation}
where $\{\cdot,\cdot\}$ denotes the process of merging the latter to the former.
If this step is successfully performed, the processing of an outlier sample $v^o_i$ for $OPLM_{v \rightarrow t} $ is completed.
If $v_i$ is an outlier sample, it will be added to the collection $U^v$. 
The process will then return to formula \cref{eq:oplm-3} for re-selection until $P^v_{filter} = U^v$. 
If the outlier sample $v^o_i$ remains unassigned to a specific category, it will be discarded, and the algorithm will proceed to the next outlier sample.
The text outlier samples are processed by the same processsing as above.

\begin{table}[t]
  \small
  \centering
  \setlength{\tabcolsep}{2pt}
  \begin{tabular}{l|ccccc}
  \toprule
  Method                   &R@1            &R@5         &R@10        &mAP         &mINP                \\
  \midrule
  \multicolumn{6}{l}{Fully Supervised Text-based Person Retrieval} \\
  \midrule
  PDG\cite{TCSVT-5}                 &69.47             &87.13          &92.13          &60.56          &-              \\
  IRRA\cite{IRRA}                   &73.38 &89.93 &93.71 &66.13 &50.24\\
  RaSa\cite{bai2023rasa}            &76.51 &90.29 &94.25 &69.38 &- \\
  APTM\cite{yang2023towards}        &76.53 & 90.04 &94.15 &66.91 &- \\
  CFAM\cite{zuo2024ufinebench}      &76.71 &91.83 &95.96 &68.47 &- \\
  Tan \etal \cite{tan2024harnessing}&76.82 &- &- &69.55 &- \\
  DualFocus\cite{deng2024dualfocus} &77.43 &90.73 &94.20 &68.35 &53.56 \\
  MARS\cite{ergasti2024mars}        &77.62 &90.63 &94.27 &71.41 &- \\ 
  BAMG\cite{cheng2024ACCV}          &\textbf{79.98} &\textbf{92.31} &\textbf{94.03} &\textbf{68.55} &- \\
  \midrule
  \multicolumn{6}{l}{Weakly Supervised Text-based Person Retrieval} \\
  \midrule
  CMMT\cite{CMMT}                   &57.10              &78.14          &85.23          &-              &- \\
  CMMT(CLIP-ViT-B/16)*              &59.57 &79.53 &86.53 &54.66 &39.78 \\
  CAIBC\cite{CAIBC}                 &58.64              &79.02          &85.93      &-              &- \\
  ECCA\cite{gong2024enhancing}      &68.13 &87.26 & 91.88 &- &- \\
  CCL\cite{du2024contrastive}       &67.25 &86.10 &91.45 &60.83 &- \\
  CANC\cite{gong2024cross}          &69.61 &88.10 &91.47 &- &- \\
  \textbf{CPCL(CLIP-RN50)}          &60.09              &80.72          &87.70      &53.47          &36.57 \\
  \textbf{CPCL(CLIP-RN101)}         & 64.88             &84.02          &89.70      &57.76          &41.22 \\
  \textbf{CPCL(CLIP-ViT-B/16)}      &\textbf{70.03}     &\textbf{87.28} &\textbf{91.78} &\textbf{63.19} &\textbf{47.54} \\
  \bottomrule
  \end{tabular}
  \caption{Performance comparisons with SOTA methods on CUHK-PEDES. 
  ``G" and ``L" correspond to the global-matching and local-matching methods, respectively. (*) denotes our implementation based on the CMMT's paper.}
  \label{tab:CUHK}
\end{table}

\subsubsection{The Supplementary Learning Stage}
At the supplementary learning stage, the image-text pairs that are not mined by the refined learning stage compose the training dataset.
The main objective loss function is ITC loss.
For each image feature $f^v_i$, we obtain an InfoNCE\cite{infoNCE} loss between its image feature $f^v_i$ and all text features $f^t_j$ in the batch,
\begin{equation} \label{eq:I2T}
\mathcal L_{v2t}^{itc} = 
        -\log
        \frac{ \exp \left(f^v_i \cdot f^{t}_i / \tau \right)}
        {\sum_{k=1}^{N} \exp \left(f^v_i \cdot f^t_{k} / \tau\right)},
\end{equation}
where the $N$ denotes the batch size, and $\tau$ denotes the hyperparameter of temperature. 
Symmetrically, for each text feature, the InfoNCE loss is denoted as $\mathcal L_{t2i}^{itc}$.
The total loss function for ITC loss is defined as:
\begin{equation} \label{eq:ITC}
\mathcal L_{itc} = 
    \mathcal L_{v2t}^{itc} + \mathcal L_{t2v}^{itc}.
\end{equation}
For convenience, we will refer to an OPLM that contains only the first stage as a \textit{One-stage OPLM} and an OPLM that contains two stages as a \textit{Two-stage OPLM}.

\section{Experiments}

\subsection{Evaluation Metrics and Dataset}
For all our experiments, we utilize recall at Rank-\textit{k} (\textit{k}=1, 5, 10) metric to evaluate the performance of the retrieval system. 
Additionally, for a comprehensive evaluation, we incorporate the mean Average Precision (mAP) and mean Inverse Negative Penalty (mINP)\cite{mINP} as additional retrieval criteria. 
For dataset, we use CUHK-PEDES, ICDG-PEDES, and RSTPReid. For more information on them, please refer to \cref{section:dataset}.

\begin{table}[t]
  \small
  \centering
  \setlength{\tabcolsep}{2pt}
  \begin{tabular}{l|ccccc}
    \toprule
    Method                             &R@1       &R@5      &R@10       &mAP        &mINP         \\
    \midrule
    \multicolumn{6}{l}{Fully Supervised Text-based Person Retrieval} \\
    \midrule
    PDG\cite{TCSVT-5}           &57.69  &75.79  &82.67  &36.07  &- \\    
    IRRA\cite{IRRA}             &63.46 &80.25 &85.82 &38.06 &7.93 \\
    RaSa\cite{bai2023rasa}      &65.28 &80.40 &85.12 &41.29 &-\\
    APTM\cite{yang2023towards} & 68.51 &82.09 &87.56 &41.22 &- \\
    CFAM\cite{zuo2024ufinebench} &66.58 &82.47 &87.37 &40.46 &- \\
    DualFocus\cite{deng2024dualfocus} &67.87 &81.93 & 87.13 &40.13 &9.14 \\
    Tan\etal\cite{tan2024harnessing} &67.05 &- &- &41.51 &- \\
    MARS\cite{ergasti2024mars} &67.60 &81.47 &85.79 &44.93 &- \\
    BAMG\cite{cheng2024ACCV} &\textbf{71.70} &\textbf{86.34} &\textbf{89.71} &\textbf{42.37} &- \\
    \midrule
    \multicolumn{6}{l}{Weakly Supervised Text-based Person Retrieval} \\
    \midrule
    ECCA\cite{gong2024enhancing} &58.89 &77.46 &83.34 &- &- \\
    CCL\cite{du2024contrastive} &58.33 &76.75 &83.38 &32.66 &- \\
    CANC\cite{gong2024cross} &60.52 &78.36 &84.13 &- &- \\
    \textbf{CPCL(CLIP-RN50)}         &52.20          &72.45          &79.77          &28.82         &4.12 \\
    \textbf{CPCL(CLIP-RN101)}        &55.15          &74.38          &81.24          &30.47         &4.47 \\ 
    \textbf{CPCL(CLIP-ViT-B/16)}     &\textbf{62.60} &\textbf{79.07} &\textbf{84.46} &\textbf{36.16} &\textbf{6.31} \\
    \bottomrule
    \end{tabular}
  \caption{Performance comparisons with methods on ICFG-PEDES dataset. (*) denotes our implementation based on the CMMT's paper}
  \label{tab:ICFG}
\end{table}

\begin{table}[t]
  \small
  \centering  
  \setlength{\tabcolsep}{2pt}
  \begin{tabular}{l|ccccc}
    \toprule
    Method                   &R@1       &R@5      &R@10       &mAP        &mINP         \\
    \midrule
    \multicolumn{6}{l}{Fully Supervised Text-based Person Retrieval} \\
    \midrule
    IRRA\cite{IRRA}          &60.20  &81.30  &88.20   &47.17     &25.28 \\
    RaSa\cite{bai2023rasa}   &66.90     &86.50     &91.35   &52.31 &- \\
    APTM\cite{yang2023towards} &67.50 &85.70 &91.45 &52.56 &- \\
    CFAM\cite{zuo2024ufinebench} &63.54 &84.75 &92.32 &50.48 &- \\
    DualFocus\cite{deng2024dualfocus} &69.12 &86.68 &92.31 &52.55 &27.87 \\
    Tan\etal\cite{tan2024harnessing} &68.50 &- &- &53.02 &- \\
    MARS\cite{ergasti2024mars} &67.55 &86.65 &91.35 &52.92 &- \\
    BAMG\cite{cheng2024ACCV} &\textbf{69.73} &\textbf{87.65} &\textbf{93.33} &\textbf{55.21} &- \\
    \midrule
    \multicolumn{6}{l}{Weakly Supervised Text-based Person Retrieval} \\
    \midrule
    CMMT(CLIP-ViT-B/16)*     &52.25      &76.45  &84.55   &41.98  &22.00 \\
    ECCA\cite{gong2024enhancing} &58.49 &80.75 &85.60 &- &- \\
    CCL\cite{du2024contrastive} &51.30 &75.25 &84.60 &41.10 &- \\
    CANC\cite{gong2024cross} &56.24 &80.15 &86.61 &- &- \\
    \textbf{CPCL(CLIP-RN50)} &49.55      &75.85   &84.15     &37.37      &16.22 \\
    \textbf{CPCL(CLIP-RN101)} &52.60      &78.30   &85.45     &39.14      &17.06 \\  
    \textbf{CPCL(CLIP-ViT-B/16)}&\textbf{58.35}      &\textbf{81.05}        &\textbf{87.65}      &\textbf{45.81}      &\textbf{23.87} \\
    \bottomrule
  \end{tabular}
  \caption{Performance comparisons with methods on RSTPReid dataset. (*) denotes our implementation based on the CMMT's paper}
  \label{tab:RSTP}
\end{table}

\subsection{Implementation Details}
Motivated by the successful use of CLIP knowledge to text-image person retrieval, we directly initialize our CPCL with the full CLIP model to enhance our method crucial cross-modal alignment capabilities. More details about the implementation, please infer to \cref{section: details}.

\subsection{Comparison with State-of-the-art Methods}
As CMMT\cite{CMMT} is only tested on the CUHK-PEDES dataset, we implemented CMMT with CLIP as the backbone to access its performance on ICFG-PEDES and RSTPeid.
We also conduct domain generalization experiments to assess the generalization ability of CPCL which are reported in the \cref{section: domain generalization} due to space limitations.

\subsubsection{Comparison with Weakly supervised Text-based Person Retrieval Methods}
We compared our method with prior methods on three datasets, as shown in \cref{tab:CUHK}, \cref{tab:ICFG} and \cref{tab:RSTP}, respectively.
Our method CPCL achieves the SOTA performance in terms of all metrics, outperforming existing methods by a large margin.
Specifically, compared with the SOTA, CPCL gains a significant Rank-1 improvement of 0.42\%, 2.08\% and 2.11\% on the three datasets, respectively.

\subsubsection{Comparison with Fully Supervised Text-based Person Retrieval Methods}
To see the gaps between weakly and fully supervised text-based person retrieval, we also list the performance of the fully supervised text-based person retrieval methods in the 1st group on three datasets, as shown in \cref{tab:CUHK}, \cref{tab:ICFG} and \cref{tab:RSTP} respectively.
Our method remains highly competitive even in comparison to fully supervised text-based person retrieval approaches, surpassing several of them in performance.

\begin{table}[t]
  \centering
  \small
  \begin{tabular}{c|ccc|ccc}
  \toprule
  \multirow{2}{*}{No.} &\multicolumn{3}{c|}{Components} &\multicolumn{3}{c}{CUHK-PEDES} \\ 
  \cmidrule{2-7} 
        &OPLM &$\mathcal{L}_{pcm}$ &$\mathcal{L}_{icpm}$    &R@1  &R@5  &R@10 \\ 
  \midrule
  0      &-         &-         &-           &58.45   &78.87   &85.30     \\
  1      &\ding{51} &-         &-           &61.44   &81.48   &88.56     \\ 
  2      &-         &\ding{51} &-           &65.14   &84.62   &90.51    \\
  3      &-         &-         &\ding{51}   &65.08   &83.38   &89.30    \\ 
  4      &-         &\ding{51} &\ding{51}   &68.76   &86.76   &91.23  \\
  5      &\ding{51} &-         &\ding{51}   &67.99   &86.08   &91.63   \\ 
  6      &\ding{51} &\ding{51} &-           &68.24   &86.06   &91.26 \\
  \midrule
  7  &\ding{51}&\ding{51}&\ding{51}  &\textbf{70.03} &\textbf{87.28} &\textbf{91.78} \\ 
  \bottomrule
  \end{tabular}%
  \caption{Ablation studies on our proposed components of CPCL.}
  \label{tab:ablation}
\end{table}

\begin{table}[t]
  \centering
  \small
  \setlength{\tabcolsep}{1mm}
  \begin{tabular}{c|ccc|ccc}
  \toprule
  \multirow{2}{*}{No.} &\multicolumn{3}{c|}{Components}  &\multicolumn{3}{c}{CUHK-PEDES} \\ 
  \cmidrule{2-7} 
    &$PCM_c$        &\makecell{Memory \\ update}       &\makecell{Two-stage \\ OPLM}    &R@1  &R@5  &R@10 \\ 
    \midrule
  0 &-     &\ding{51}    &\ding{51}    &68.24  &86.06   &91.23 \\
  1 &\ding{51}      &-   &\ding{51}    &69.10  &86.18   &91.19 \\ 
  2 &\ding{51}      &\ding{51}    &-   &68.76  &86.11   &91.26 \\
  \midrule
  3 &\ding{51} &\ding{51} &\ding{51} &\textbf{70.03}  &\textbf{87.28} &\textbf{91.78} \\
  \bottomrule
  \end{tabular}%
  \caption{Ablation studies on optional components of CPCL.}
  \label{tab:ablation-others}
\end{table}

\subsection{Ablation Study}
In this section, we thoroughly analyze the effectiveness of each module in the CPCL.
The overall results are reported in \cref{tab:ablation}.

\subsubsection{Prototypical Multi-modal Memory (PMM) and Hybrid-level Cross-modal Matching (HCM)}
In contrast to existing CLIP-based methods that solely establish one-to-one connections between image-text pairs, 
the proposed HCM module establishes the intricate associations within heterogeneous modalities of image-text pairs of the same person through HCM.
It should be noted that PMM and HCM need to be combined to work properly.
HCM comprises two components: PCM and ICPM.
As shown in \cref{tab:ablation}, No.0 \vs No.2, No.0 \vs No.3, and No.1 \vs No.6, we readily observe that the inclusion of the PMM and HCM proves beneficial for enhancing the model's performance. 
Removing this module will result in a certain degree of performance degradation.

\subsubsection{Outlier Pseudo Label Mining (OPLM)}
As shown in \cref{tab:ablation}, No.0 \vs No.1,  No.2 \vs No.6 and No.3 \vs No.5, we conducted an ablation study of OPLM. 
OPLM makes significant contributions to our method's performance. 
Being a plug-and-play component, OPLM can seamlessly integrate into weakly supervised text-based person retrieval methods. 
These findings unequivocally demonstrate the effectiveness and flexibility of our OPLM approach. 
The removal of this module results in the inability to extract valuable diverse samples, leading to performance degradation of the model.
We record the number of outliers during training and OPLM effectively reduces the number of outliers. Please refer to \cref{section: quantity of outliers} for more detailed analysis.

\subsubsection{cross-modal PCM \vs single-modal PCM}
We tried two types of prototypical contrast matching (PCM): \textit{cross-modal PCM} and \textit{single-modal PCM}, denoted as $PCM_c$ and $PCM_s$. In our all experiments, the default is $PCM_c$.
The experimental results are shown No.0 \vs No.3 in \cref{tab:ablation-others} and
$PCM_s$ achieved a competitive performance on CUHK-PEDES.
But $PCM_c$ surpassed $PCM_s$ in terms of all metrics and get better performance on CUHK-PEDES.
$PCM_c$ incorporates both intra-modal and inter-modal prototype-based contrastive matching losses and enables more comprehensive representation learning across and within modalities, which is verified by visualization of t-SNE\cref{section: qualitative results}.

\subsubsection{Two-stage OPLM \vs One-stage OPLM}
Although we have developed a One-stage OPLM capable of mining valuable unclustered samples, certain image-text pairs remain unminable. 
Recognizing this limitation, we introduce a Two-stage OPLM to fully leverage these unminable image-text pairs. 
We conduct experiments to assess the effectiveness of this two-stage training scheme. 
As evidenced by the comparison between No.3 \vs No.2 in \cref{tab:ablation-others}, CPCL demonstrates superior performance when trained with the Two-stage OPLM scheme.

\subsubsection{Exploration of Momentum updating scheme for PMM}
PMM adopts a momentum updating scheme to dynamically refine the prototypical features.
The performance gap between No.1 and No.3 in \cref{tab:ablation-others} demonstrates the importance of this scheme and its positive effect on CPCL.
The dynamic momentum updating scheme offers the advantage of providing more confident learning targets. However, this approach also carries inherent risks, such as the potential for high variance and instability in the training process. Interestingly, during our experiments, we obviously did not observe these adverse phenomena. This suggests that the benefits of dynamic updating may outweigh its risks in certain contexts.

\begin{table}[t]
  \centering
  \small
  \begin{tabular}{c|l|ccc}
  \toprule
  \multirow{2}{*}{No.} &\multirow{2}{*}{Methods} &\multicolumn{3}{c}{CUHK-PEDES} \\ 
  \cmidrule{3-5} 
    &                   &R@1  &R@5  &R@10 \\ \midrule
  0 &CPCL+Kmeans-5000   &68.68 &86.26 &91.16 \\
  1 &CPCL+Kmeans-8000   &69.80 &86.36 &91.11 \\ 
  2 &CPCL+Kmeans-11000  &\textbf{70.16} &86.84 &\textbf{91.89}  \\
  3 &CPCL+Kmeans-14000  &69.86 &86.24 &91.20 \\
  \midrule
  5 &CPCL(DBSCAN)       &70.03   &\textbf{87.28}   &91.78 \\
  \bottomrule
  \end{tabular}%
  \caption{Evaluate CPCL with other clustering algorithms on CUHK-PEDES.}
  \label{tab:ablation-cluster}
\end{table}

\subsubsection{Analysis of clustering algorithm}
In order to verify that CPCL is still effective with other clustering algorithms, we conduct experiments by replacing the DBSCAN algorithm with $K$-means clustering algorithms where the $K$ is set as 5000, 8000, 11000 and 14000 for CUHK-PEDES.
As shown in \cref{tab:ablation-cluster}, we observe that the $K$ value impacts the model performance. 
The closer the $K$ value is to the actual identity number, the better the performance is as shown in No.0, No.1, and No.2 in \cref{tab:ablation-cluster}.
The performance of CPCL+Kmeans-11000 (No.2) in \cref{tab:ablation-cluster} is better than that of CPCL(DBSCAN), and we attribute this result to the fact that the $K=11000$ almost equals to the actual identity number of CUHK-PEDES(11003).

\begin{table}[t]
  \centering
  \small
  \begin{tabular}{c|l|ccc}
  \toprule
  \multirow{2}{*}{No.} &\multirow{2}{*}{Initialization policy} &\multicolumn{3}{c}{CUHK-PEDES} \\ 
  \cmidrule{3-5} 
       &        &R@1  &R@5  &R@10 \\ 
  \midrule
  0    &Random  &69.14   &85.92   &91.03     \\
  1    &Average    &\textbf{70.03}   &\textbf{87.28}   &\textbf{91.78}    \\ 
  \bottomrule
  \end{tabular}%
  \caption{Ablation study on PCM initialization policy.}
  \label{tab:PCM-init}
\end{table}

\subsubsection{Analysis of Initialization of PMM}
In \cref{tab:PCM-init}, we evaluate different initialization policies Of PCM.
Specifically, we explored two initialization methods: ``Random Initialization" and ``Average Initialization." In the ``Random Initialization" strategy, we randomly select the feature of an instance belonging to a person to initialize his/her prototypical feature. This approach is straightforward and simple, but it is susceptible to the randomness of a single instance's feature, especially when the dataset contains noisy labels. The randomly chosen feature may not accurately represent the person's overall characteristics.
In contrast, the ``Average Initialization" strategy involves averaging the features of all samples belonging to the same person to initialize his/her prototypical feature. 
This method has the advantage of better reflecting a person's overall characteristics, thereby enhancing the robustness and reliability of the initialized feature.

\section{Conclusion}
In this paper, we propose CPCL, which demonstrates competitive performance by enhancing alignment through intra-modal and inter-modal prototype-based contrastive matching losses.
CPCL not only advances the field of weakly supervised text-based person retrieval but also offers a promising foundation for future research in pseudo-label refinement and domain generalization.
We hope that our work can inspire and facilitate further innovations in weakly supervised learning and multi-modal data integration.
Despite the encouraging results, a noticeable performance gap remains when compared to fully supervised text-based person retrieval.
We believe that the quality of pseudo labels is a critical factor contributing to this gap.
Future work will focus on improving pseudo-label quality and exploring more robust alignment strategies to further narrow this gap.

\bibliography{aaai2026}

\newpage
\appendix

\section{Appendix}
In this appendix, we present additional experimental results and detailed analyses that could not be included in the main paper due to space limitations.
\cref{section: qualitative results} presents the qualitative results, such as the visualization of embeddings of images and text, visualization of Saliency map, and visulization of retrieval results.
\cref{section: OPLM} shows the number of outliers during trianing, which demonstrates the effectiveness of our Outlier Pseudo Label Mining (OPLM) module.
\cref{section: domain generalization} offers the performance of our method in the domain generalization problem.
Finally, \cref{section: clustering algo} delves into the ifluence of clustering hyper-parameters on our method.
\cref{section: details} provides the implementations details in our experiments.
\cref{section:dataset} provides the details of CUHK-PEDES, ICFG-PEDES, and RSTPReid datasets.

\begin{figure}[h]
    \centering
    \includegraphics[scale=0.9]{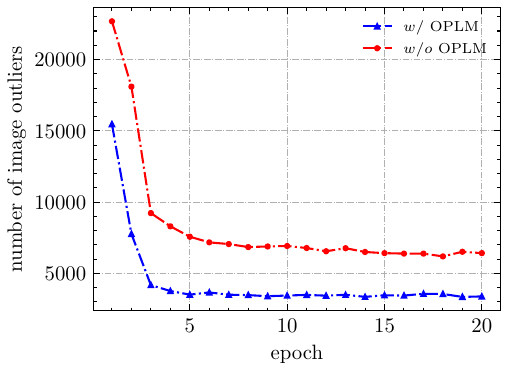}
    \caption{Illustration of the dynamically changing image outlier number on CUHK-PEDES dataset.}
    \label{fig:img-outlier}
\end{figure}

\begin{figure}[h]
    \centering
    \includegraphics[scale=0.9]{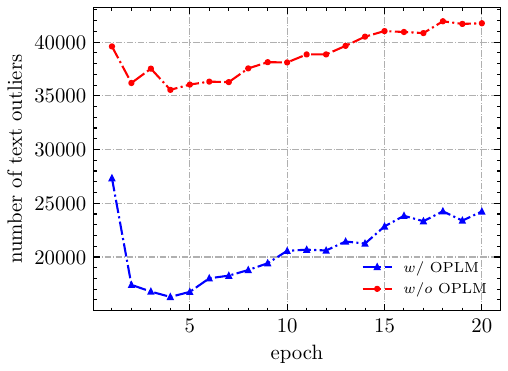}
    \caption{Illustration of the dynamically changing text outlier number on CUHK-PEDES dataset}
    \label{fig:txt-outlier}
\end{figure}

\section{The Number of Outliers during Training}\label{section: quantity of outliers}
To maximize the utility of existing image-text pairing information, we introduce an Outlier Pseudo Label Mining (OPLM) module. This module is designed to further distinguish valuable outlier samples from each modality by mining implicit relationships between image-text pairs, thereby creating more reliable clusters.

We illustrate the quantity of image and text outliers during the initial 20 epochs of training on the CUHK-PEDES dataset in \cref{fig:img-outlier} and \cref{fig:txt-outlier}. The figures clearly show that the quantity of image and text outliers is significantly reduced when the OPLM module is employed, compared to the scenario without the OPLM module.

This reduction in outlier quantity demonstrates the efficacy of our OPLM in identifying and leveraging valuable outlier samples through the implicit relationships between image-text pairs. By effectively mining these relationships, the OPLM module enhances the clustering process, leading to more accurate and reliable clusters.

\begin{figure*}[t]
    \centering
    \includegraphics[scale=1.0]{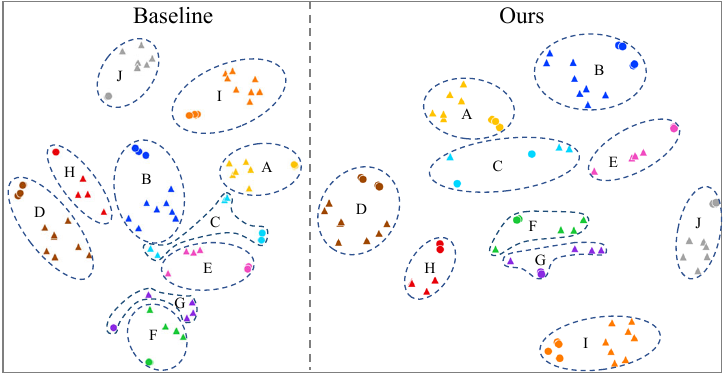}
    \caption{t-SNE visualization of random 10 person in CUHK-PEDES.
    Text and image samples are denoted as triangles and circles, respectively.
    Samples with the same color indicate that they belong to the same person.
    The same letter means that it corresponds to the same person.
    }
    \label{fig:t-sne-map}
\end{figure*}

\begin{figure}[t]
    \centering
    \includegraphics[scale=1.0]{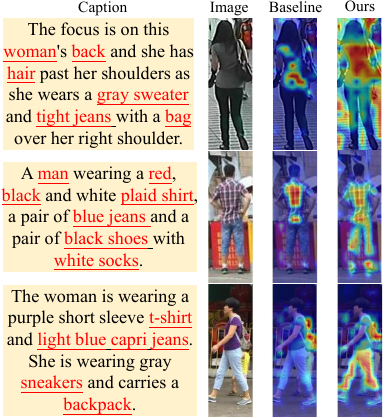}
    \caption{Visualization of Saliency Map based on Grad-CAM.
    The samples are from CUHK-PEDES.}
    \label{fig:saliency-map}
\end{figure}

\begin{figure}[t]
    \centering
    \includegraphics[scale=0.9]{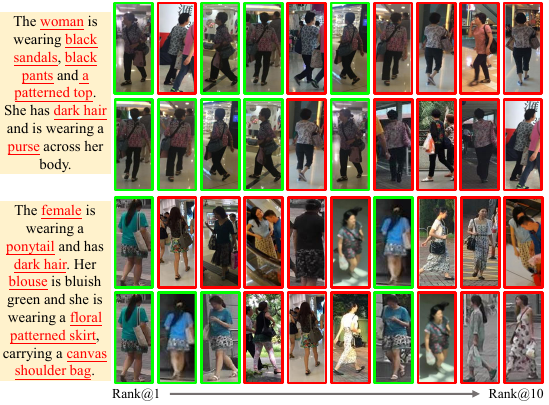}
    \caption{Comparison of the top-10 retrieved results on the CUHK-PEDES using the baseline (first row) and CPCL method (second row).
    The images corresponding to the query text, as well as the matched and mismatched images, are distinguished by green and red rectangles, respectively.}
    \label{fig:retrieval}
\end{figure}
\section{Analysis of Qualitative Results}\label{section: qualitative results}
\subsection{Visualization of t-SNE}
We employ t-SNE\cite{van2008visualizing} to visualize the embeddings of image and text samples from 10 randomly selected individuals in the CUHK-PEDES dataset. As depicted in \cref{fig:t-sne-map}, our approach demonstrates superior intra-class compactness and inter-class separability compared to the baseline. This enhanced compactness and separability are instrumental in achieving better weakly supervised text-based person retrieval performance.

The improved intra-class compactness ensures that samples from the same person are clustered more tightly, while the enhanced inter-class separability ensures that samples from different individuals are more distinctly separated. This dual improvement not only facilitates the model's ability to distinguish between different individuals but also enhances its robustness in recognizing individuals under weak supervision.

Additionally, our analysis reveals that text samples exhibit greater diversity compared to image samples. This observation aligns with the inherent nature of the data, where annotators often describe the same image in various ways. This diversity in textual descriptions suggests that the challenge of weakly supervised text-based person retrieval may lie in effectively processing and leveraging these varied text samples.

\subsection{Visualization of Saliency Map}
To gain deeper insights into the alignment between visual and textual modalities, we employ Grad-CAM\cite{grad-CAM} to compute the saliency map based on the [CLS] token of the last attention layer in the image encoder. The objective of this analysis is to measure the cosine similarity between the encoding of the input image and the text encoding.

As illustrated in \cref{fig:saliency-map}, our method effectively guides the model to learn the semantic relevance between visual and textual modalities. This is evident in the model's ability to focus more on the attributes of human bodies described by the text, as compared to the baseline. This enhanced focus on textual attributes suggests that our approach not only aligns the visual and textual data but also enriches the model's understanding of the semantic context provided by the text.

Furthermore, this improved alignment can be attributed to the model's ability to capture nuanced relationships between visual features and textual descriptions. By leveraging the saliency maps, we can observe how the model prioritizes certain visual elements based on the textual input, which is crucial for tasks that require a high degree of semantic understanding, such as image captioning or visual question answering.

\subsection{Visualization of retrieval results}
We present the top-10 retrieval results obtained using textual descriptions from both the baseline and CPCL (Cross-Prototype Classification Learning). As illustrated in \cref{fig:retrieval}, CPCL demonstrates superior accuracy in retrieving relevant images for a given textual query. This improvement is attributed to the model's enhanced ability to align visual and textual modalities more effectively.

However, it is important to note that our model also returned an incorrect retrieval result. This occurrence can be attributed to the global matching method employed in our model, which directly aligns global visual and textual embeddings. While this approach offers computational efficiency, it may inadvertently overlook some crucial local details that are essential for accurate retrieval.

This observation highlights a potential limitation of our current model and suggests that future work should focus on refining the matching strategy. One promising direction is to incorporate local feature matching, which could provide a more nuanced understanding of the visual and textual content. By combining global and local information, we aim to enhance the model's ability to capture fine-grained details, thereby improving overall retrieval accuracy.

\begin{table}[h]
\centering
\small
\begin{tabular}{c|l|ccc}
\toprule
\multirow{2}{*}{No.} &\multirow{2}{*}{Methods}  &\multicolumn{3}{c}{CUHK $\rightarrow$ ICFG} \\
\cmidrule{3-5}
 & & R@1 & R@5 & R@10 \\
\midrule
\multicolumn{5}{l}{Fully Supervised Text-based Person Retrieval} \\
\midrule
1 & MIA \cite{niu2020improving}       & 19.35     & 36.78      & 46.42          \\
2 & SCAN \cite{lee2018stacked}        & 21.27     & 39.26      & 48.83            \\
3 & SSAN \cite{ICFG-PEDES}  & 29.24     & 49.00      & 58.53        \\
4 & LGUR \cite{shao2022learning}      & 34.25     & 52.58      & 60.85        \\
5 &IRRA\cite{IRRA}                    &42.40     &62.10   &69.61 \\
6 & RaSa\cite{bai2023rasa}            & \textbf{50.59} & \textbf{67.46} & \textbf{74.09}  \\
\midrule
\multicolumn{5}{l}{Weakly Supervised Text-based Person Retrieval} \\
\midrule
7 &CMMT\cite{CMMT} &31.16 &50.02 &17.31 \\
8 & CPCL(Ours) &\textbf{39.66} &\textbf{59.01} &\textbf{67.67} \\
\bottomrule
\end{tabular}
\caption{Comparison with other methods on domain generalization task. We adopt CUHK-PEDES and ICFG-PEDES as the source domain and the target domain in turn.}
\label{tab:DG-cuhk-icfg}
\end{table}

\begin{table}[h]
\centering
\small
\resizebox{\columnwidth}{!}{
\begin{tabular}{c|l|ccc}
\toprule
\multirow{2}{*}{No.} &\multirow{2}{*}{Methods}  &\multicolumn{3}{c}{ICFG $\rightarrow$ CUHK} \\
\cmidrule{3-5}
 & & R@1 & R@5 & R@10 \\
\midrule  
\multicolumn{5}{l}{Fully Supervised Text-based Person Retrieval} \\
\midrule
0 & Dual Path \cite{zheng2020dual}    & 7.63     & 17.14      & 23.52       \\
1 & MIA \cite{niu2020improving}       & 10.93     & 23.77      & 32.39       \\
2 & SCAN \cite{lee2018stacked}        & 13.63     & 28.61      & 37.05       \\
3 & SSAN \cite{ICFG-PEDES}  & 21.07     & 38.94      & 48.54       \\
4 & LGUR \cite{shao2022learning}      & 25.44     & 44.48      & 54.39       \\
5 & IRRA\cite{IRRA}             &33.45 &56.29 &66.31 \\
6 & RaSa\cite{bai2023rasa}      & \textbf{50.70} & \textbf{72.40} & \textbf{79.58}  \\
\midrule
\multicolumn{5}{l}{Weakly Supervised Text-based Person Retrieval} \\
\midrule
7 & CMMT\cite{CMMT} &33.07 &54.24 &63.99 \\
8 & CPCL(Ours) &\textbf{35.48} &\textbf{57.52} &\textbf{66.57} \\
\bottomrule
\end{tabular}}
\caption{Comparison with other methods on domain generalization task. We adopt ICFG-PEDES and CUHK-PEDES as the source domain and the target domain in turn.}
\label{tab:DG-icfg-cuhk}
\end{table}
\section{Domain Generalization}\label{section: domain generalization}
We perform a series of domain generalization experiments to assess the generalization ability of CPCL. Specifically, we evaluate the model's performance on a target domain using a model trained on a source domain, as detailed in \cref{tab:DG-cuhk-icfg} and \cref{tab:DG-icfg-cuhk}.

These experiments are designed to test how well CPCL can adapt to new, unseen domains without additional training. By transferring the model from a source domain to a target domain, we aim to understand the extent to which CPCL can generalize its learned representations across different datasets.
The experiments demonstrate that our model achieves commendable performance. While it does not yet match the performance of state-of-the-art supervised methods, we are pleasantly surprised to find that our approach outperforms many existing supervised methods.
The fact that our weakly supervised model can surpass several supervised methods suggests that the underlying principles and techniques we have developed are highly effective.

This finding also raises intriguing questions about the potential of weakly supervised learning in domains traditionally dominated by fully supervised approaches. 
There are three possible answers to this phenomenon: one is that supervised methods suffer from overfitting problems, two is that the labels in the training data are noisy, and three is that both exist.
Since this problem is beyond the scope of this paper, we intend to explore the boundaries of this problem in subsequent work.

Our domain generalization experiments not only validate the robustness of CPCL but also open up new research directions for enhancing cross-domain adaptability, which is essential for deploying machine learning models in real-world scenarios.

\section{Impact of Clustering Hyper-parameters}\label{section: clustering algo}
We use DBSCAN\cite{DBSCAN} algorithm and Jaccard distance\cite{zhong2017re} with $k$-reciprocal nearest neighbors for clustering images and texts separately, where the $k=20$.
As for images, the maximum distance $d_v$ between neighbors is set as 0.5 and the minimal number of neighbors $neb_v$ for a dense point is set as 2.
As for texts, the maximum distance $d_t$ between neighbors is set as 0.6 and the minimal number of neighbors $neb_t$ for a dense point is set as 4.

The maximum distance and minimal number of neighbors for a dense point are hyper-parameters of the DBSCAN algorithm, which affect the final number of clusters.
In general, if the $d_v$ and $d_t$ is chosen too large, the clusters will merge and it is possible that a cluster will contain a very large number of samples.
Conversely, if they are set very small, then most of the samples will be considered as outliers.
As for the minimal number of neighbors $neb_v$ and $neb_t$, the larger they are, the more samples will be considered as outliers. 
Conversely, if the value of they is too small, then there will be a number of core points, which in turn will lead to multiple clusters being merged into one.

We analyze the influence of them on the CUHK-PEDES as shown in \cref{tab:abalition-dbscan-img} and \cref{tab:abalition-dbscan-txt}.
We find that the image samples are more sensitive to different hyper-parameters, as No.2 and No.5 shown in \cref{tab:abalition-dbscan-img}.
On the contrary, the text samples are more robust to different hyper-parameters, and the performance of the OPLM fluctuates very little for different parameters.
We conjecture that this is related to the number of samples in the different modalities.
As for the CUHK-PEDES dataset, each image will typically correspond to two text samples, so the number of samples for text is approximately twice as large as for images.

\begin{table}[h]
  \centering
  \small
  \begin{tabular}{c|c|c|ccc}
  \toprule
  \multirow{2}{*}{No.} &\multirow{2}{*}{$d_v$} &\multirow{2}{*}{$neb_v$} &\multicolumn{3}{c}{CUHK-PEDES} \\ 
  \cmidrule{4-6} 
       &       &        &R@1  &R@5  &R@10 \\ \midrule
  0  &0.3  &\multirow{3}{*}{2}  &\textbf{70.31}   &86.84   &\textbf{92.00}     \\
  1    &0.5    &        &70.03  &\textbf{87.28}   &91.78    \\
  2    &0.7    &        &60.58  &79.18   &86.08 \\
  \midrule
  3 &\multirow{3}{*}{0.5} &2 &\textbf{70.03} &\textbf{87.28} &\textbf{91.78} \\
  4 & &4    &69.51 &87.09 &91.76 \\
  5 & &8    &65.03 &83.92 &89.98 \\
  \bottomrule
  \end{tabular}%
  \caption{Performances of CPCL with different of $d_v$ and $neb_v$, where $d_v$ represents the maximum distance between two image samples and $neb_v$ represents the minimal number of image neighbors for an image dense point in DBSCAN. $d_t = 0.6$ and $neb_t=4$.}
  \label{tab:abalition-dbscan-img}
\end{table}

\begin{table}[h]
  \centering
  \small
  \begin{tabular}{c|c|c|ccc}
  \toprule
  \multirow{2}{*}{No.} &\multirow{2}{*}{$d_t$} &\multirow{2}{*}{$neb_t$} &\multicolumn{3}{c}{CUHK-PEDES} \\ 
  \cmidrule{4-6} 
       &       &        &R@1  &R@5  &R@10 \\ \midrule
  0  &0.4  &\multirow{3}{*}{4}  &68.97   &86.06   &91.16     \\
  1    &0.6    &        &\textbf{70.03} &\textbf{87.28} &\textbf{91.78}    \\
  2    &0.8    &        &69.75 &86.34  &91.47 \\
  \midrule
  3 &\multirow{3}{*}{0.6} &2 &70.04 &86.40 &\textbf{91.99} \\
  4 & &4    &70.03   &\textbf{87.28}   &91.78 \\
  5 & &8    &\textbf{70.42} &86.55 &91.78 \\
  \bottomrule
  \end{tabular}%
  \caption{Performances of CPCL with different of $d_t$ and $neb_t$, where $d_t$ represents the maximum distance between two text samples and $neb_t$ represents the minimal number of text neighbors for a text dense point in DBSCAN. $d_v = 0.5$ and $neb_v=2$.}
  \label{tab:abalition-dbscan-txt}
\end{table}

\section{Implementation Details}\label{section: details}
\subsubsection{The Hyper-parameters in training}
The batch size is set to 128, and the model is trained with Adam optimizer\cite{adam} for 60 epochs with a learning rate initialized to $1 \times 10^{-5}$ and cosine learning rate decay.
In the beginning, the 5 warm-up epochs linearly increase the learning rate from $1 \times 10^{-6}$ to $1 \times 10^{-5}$.
Following the clustering-based method CMMT\cite{CMMT}, we use DBSCAN\cite{DBSCAN} for clustering before each epoch.
All experiments are performed on a single RTX3090 24GB GPU.

\textbf{The Text Encoder of CLIP.}
In this study, we adopt a CLIP text encoder to extract the text representation, which is a Transformer\cite{Transformer} model that has been modified by Radford \etal \cite{CLIP}. 
Following the CLIP, we first employ lower-cased byte pair encoding (BPE) with a vocabulary size of 49152\cite{sennrich2015neural} to tokenize the input text description. 
To demarcate the beginning and end of the sequence, we enclose the tokenized text $\{f^t_{sos}, f^t_1, \ldots, f^t_{eos}\}$ with special [SOS] and [EOS] tokens. 
Subsequently, these tokenized text representations are fed into the transformer and capture correlations among each path by masked self-attention. 
Finally, we obtain the global text representation by linearly projecting the last layer of the transformer at the [EOS] token $f^t _{eos}$ into the image-text joint embedding space.

\textbf{The Image Encoder of CLIP.}
CLIP image encoder is a ViT\cite{ViT} model. Given an input image $I \in R^{H \times W \times C}$, where H, W, and C denote its height, width, and number of channels, respectively. 
In order to input Image $I$, we first split $I$ into $N=H \times W / P^2$ fixed-sized patches $\{f^v_i|i = 1, 2, · · ·, N \} \in R^{1 \times D}$, where N, P and D denote number, size and dim of the patch after a trainable linear projection.
Spatial information is incorporated by adding learnable position embeddings and an extra learnable [CLS] embedding token denoted as $I_{cls}$ is prepended to the input patch token sequences $\{f^v_{cls}, f^v_1, \ldots, f^v_N \}$.
Then the patch token sequences are input into L-layer transformer blocks to model correlations of each patch token.
Finally, a linear projection is adopted to map $f^v_{cls}$ to the joint image-text embedding space, which serves as a global image representation.

\section{Details of Dataset}\label{section:dataset}
\subsection{CUHK-PEDES}
The CUHK-PEDES\cite{CUHK-PEDES} dataset represents a pioneering effort in text-to-image person retrieval, encompassing 40,206 images and 80,412 textual descriptions associated with 13,003 individual identities. 
The training set comprises 11,003 identities, 34,054 images, and 68,108 textual descriptions, while the validation and test set each includes 3,078 images and 6,158 textual descriptions, respectively, representing a total of 1,000 unique identities in each set.

\subsection{ICFG-PEDES}
ICFG-PEDES\cite{ICFG-PEDES} is a dataset proposed in 2021, built upon the MSMT17\cite{MSMT17} dataset. 
It is unique in that each pedestrian image is associated with only one text description. 
The dataset consists of 54,522 pedestrian images and an equal number of fine-grained text descriptions, covering a total of 4,102 individual pedestrians. 
As per official settings, the data is divided into a training set and a test set. The training set comprises 34,674 image-text pairs, corresponding to 3,102 pedestrian IDs, while the test set contains 19,848 image-text pairs, involving 1,000 pedestrian IDs. 
While this dataset has a smaller number of text descriptions compared to the CUHK-PEDES dataset, the descriptions provided are more detailed and fine-grained.

\subsection{RSTPReid}
The RSTPReid\cite{RSTPReid} dataset, comprises 20,505 images and 41,010 corresponding text descriptions, relating to a total of 4,101 unique pedestrians. 
The images were captured from 15 cameras, enhancing the dataset's realism by replicating real-world scenarios. 
Notably, this dataset guarantees that each image is associated with precisely two text descriptions, and each pedestrian is depicted by five images captured from multiple cameras. 
The dataset is thoughtfully divided into training, verification, and test sets, as per official guidelines. 
The training set comprises 18,505 images, 37,010 text descriptions and 3,701 individual pedestrians. 
Both the test and training sets include 1,000 images and 2,000 descriptions, collectively representing 200 distinct pedestrians.

\end{document}